\documentclass[fleqn, 10pt]{wlscirep} %
\usepackage[utf8]{inputenc}
\usepackage[T1]{fontenc}
\usepackage{lineno}
\usepackage{todonotes}
\usepackage{gensymb}
\usepackage{rotating}
\usepackage[acronym]{glossaries}
\usepackage{fontawesome}
\usepackage{caption}
\usepackage{subcaption}

\newacronym{dl}{DL}{deep learning}
\newacronym{dnn}{DNN}{deep neural network}
\newacronym{ml}{ML}{machine learning}
\newacronym{casa}{CASA}{computer-assisted semen analysis}

\newcommand{\numberofvideos}{20}
\newcommand{\numberofsplitsrest}{166}
\newcommand{\numberofsplitsunlabelled}{336}

\title{VISEM-Tracking, a human spermatozoa tracking dataset}

\author[1,$\dag$]{Vajira Thambawita}
\author[1,$\dag$]{Steven A. Hicks}
\author[1,2,$\dag$]{Andrea M. Stor{\aa}s}
\author[1]{Thu Nguyen}
\author[2]{\mbox{Jorunn M. Andersen}}
\author[2]{Oliwia Witczak}
\author[2]{Trine B. Haugen}
\author[2,1,$\dag$]{Hugo L. Hammer}
\author[1,2]{\mbox{P{\aa}l Halvorsen}}
\author[1,2,$\dag$]{Michael A. Riegler}

\affil[1]{SimulaMet, Oslo, Norway}
\affil[2]{OsloMet, Oslo, Norway}

\affil[*]{corresponding author(s): Vajira Thambawita (vajira@simula.no)}
\affil[$\dag$]{these authors contributed equally to this work}

\begin{abstract} 
A manual assessment of sperm motility requires microscopy observation, which is challenging due to the fast-moving spermatozoa in the field of view. To obtain correct results, manual evaluation requires extensive training. Therefore, computer-assisted sperm analysis (CASA) has become increasingly used in clinics. Despite this, more data is needed to train supervised machine learning approaches in order to improve accuracy and reliability in the assessment of sperm motility and kinematics. In this regard, we provide a dataset called VISEM-Tracking with $20$ video recordings of $30$ seconds (comprising $29,196$ frames) of wet sperm preparations with manually annotated bounding-box coordinates and a set of sperm characteristics analyzed by experts in the domain. In addition to the annotated data, we provide unlabeled video clips for easy-to-use access and analysis of the data via methods such as self- or unsupervised learning. As part of this paper, we present baseline sperm detection performances using the YOLOv5 deep learning (DL) model trained on the VISEM-Tracking dataset. As a result, we show that the dataset can be used to train complex DL models to analyze spermatozoa.    
\end{abstract}

\glsresetall

\begin{document}

\flushbottom
\maketitle

\thispagestyle{empty}

\section*{Background \& Summary}

\Gls{ml} is increasingly being used to analyze videos of spermatozoa under a microscope for developing computer-assisted sperm analysis (CASA) systems \cite{gill2022looking, riegler2021artificial}. In the last few years, several studies have investigated the use of \glspl{dnn} to automatically determine specific attributes of a semen sample, like predicting the proportion of progressive, non-progressive, and immotile spermatozoa~\cite{thambawita2019stacked, hicks2019machine, thambawita2019extracting, javadi2019novel, you2021machine}. However, a major challenge with using \gls{ml} for semen analysis is the general lack of data for training and validation. Only a few open labeled datasets exist (Table~\ref{table:Overview_Of_Existing_Datasets}), with most focus on still-frames of fixed and stained spermatozoa or very short sequences of sperm to analyze the morphology of the spermatozoa.

\begin{table}[!t]
\renewcommand{\arraystretch}{1.4}
	\centering
	\caption{Overview of existing sperm datasets.}\label{table:Overview_Of_Existing_Datasets}
	\begin{tabular}{l p{2.5cm} p{2cm} c p{1.5cm} p{5cm}} 
		\toprule
        Author   & Name/Title            &  Ground truth   & \# Images & \# Videos & Summary \\ \midrule

        Ghasemian et al.~\cite{GHASEMIAN2015409} & HSMA-DS: Human Sperm Morphology Analysis DataSet & Classification & $1,457$ & - & This dataset is for morphology analysis and the dataset consists of captured sperm cells with $\times400$ and $\times600$ magnification. \\
        
        Javadi et al.~\cite{JAVADI2019182} & MHSMA: Modified Human Sperm Morphology Analysis Dataset &  Classification & $1,540$   & -  & The dataset is for morphology analysis. This dataset consiss of only sperm heads cropped from different samples collected from 235 participants.\\
        Shaker et al.~\cite{SHAKER2017181} & HuSHeM: Human Sperm Head Morphology &  Classification & $216$     & -  &  HuSHem is for sperm morphology classification. Semen smears were fixed and stained. Contain sperm head images of $131 \times 131$ pixels. Four classes: normal, tapered, pyriform, and amorphous. \\

        Violeta et al.~\cite{chang2017gold} & SCIAN-MorphoSpermGS & Classification & $1854$ & - & This dataset is for sperm morphology analysis. The data was classified into five classes: normal, tapered, pyriform, small, and amorphous. \\
        
        Ilhan et al.\cite{Ilhan2020}  &  SMIDS: Sperm Morphology Image Data Set  &  Classification & $3,000$  & - & For morphology analysis. The data was collected from 17 subjects. The dataset has manually annotated two classes: normal and abnormal, and an automatically extracted class: Non-sperm.\\
        McCallum et al.~\cite{McCallum2019}  & - &  Classification & $1,064$   & - &  Bright-field sperm cell images are $150 \times 150$ pixels and cropped from images of six healthy donors.\\
        Chen et al.~\cite{chen2022_SVIA} & SVIA: Sperm Videos and Images Analysis dataset  &  Detection, segmentation and classification & $4,041$   & $101$ (1-2 seconds)  & The dataset consists of $278,000$ annotated objects under three subsets. The data can be used for object detection ($125,000$ annotations), segmentation ($26,000$ annotations), and classification ($125,880$ cropped objects from the images).\\
        Haugen et al.~\cite{haugen2019visemDataset} & VISEM & Regression & - & $85$ & The dataset consists of $85$ videos of $640 \times 480$ at $50$ FPS. The ground truth files have manually assessed semen analysis data, fatty acids, sex hormones, and participant-related data.  \\

        \textbf{Ours}~\cite{vajira_thambawita_2022_7293726} & \textbf{VISEM-Tracking} & \textbf{Detection, tracking, and regression} & \textbf{$29,196$ }& \textbf{$20$} (30 seconds)  & \textbf{Our dataset contain $656,334$ annotated objects with tracking details. More details about our dataset is discussed below.} \\
        \bottomrule
    \end{tabular}
\end{table}

In this paper, we present a multi-modal dataset containing videos of spermatozoa with the corresponding manually annotated bounding boxes (localization) and additional clinical information about the sperm providers from the original study~\cite{haugen2019visemDataset}. This dataset is an extension of our previously published dataset VISEM~\cite{haugen2019visemDataset}, which included videos of spermatozoa labeled with quality metrics following the World Health Organization (WHO) recommendations~\cite{world2010laboratory}. 

There have been several datasets related to spermatozoa as follows.
For example, Ghasemian et al.~\cite{GHASEMIAN2015409} have published an open sperm dataset called HSMA-DS: Human Sperm Morphology Analysis DataSet with normal and abnormal sperm cells. Experts annotated different features, namely vacuole, tail, midpiece, and head abnormality. The availability of abnormalities of these features were marked using binary notations such as $1$ or $0$, $1$ is for abnormal, and $0$ for normal.  In total, there are $1,457$ sperm cells for morphology analysis. These sperm cell images were captured with $\times400$ and $\times600$ magnification. The Modified Human Sperm Morphology Analysis Dataset (MHSMA)~\cite{JAVADI2019182} consists of $1,540$ cropped images from the HSMA-DS dataset~\cite{GHASEMIAN2015409}. This dataset was collected for analyzing different parts of sperm cells (morphology). The maximum image size in the dataset is $128 \times 128$ pixels. 

The HuSHEM~\cite{SHAKER2017181} and SCIAN-MorphoSpermGS~\cite{chang2017gold} datasets consist of images of sperm heads captured from fixed and stained semen smears. The main purpose of these datasets is sperm morphology classification into five categories, namely normal, tapered, pyriform, small, and amorphous. SMIDS~\cite{Ilhan2020} is another dataset consisting of $3000$ images cropped from $200$ stained ocular images from $17$ subjects between $19-39$ years. From $3000$ images, $2027$ patches were manually annotated as normal and abnormal. Another $973$ samples were classified as non-sperm using spatial-based automated features. McCallum et al.~\cite{McCallum2019} have published another similar dataset with bright-field sperm cells of six healthy participants within $1064$ cropped images. The main purpose of this dataset is to find correlations between sperm cells and DNA quality. However, these datasets do not provide spermatozoa's motility and kinetics features. 

Chen et al.~\cite{chen2022_SVIA} introduced a sperm dataset called SVIA (Sperm Videos and Images Analysis dataset), which contains $101$ short $1$ to $3$ seconds video clips and corresponding manually annotated objects. The dataset is divided into three subsets, namely subset-A, B, and C. Subset-A contains $101$ video clips ($30$ FPS) containing $125,000$ object locations and corresponding categories. Subset-B contains $10$ videos with $451$ ground truth segmentation masks and subset-C consists of cropped sperms for classification into $2$ categories (impurity images and sperm images). The provided video clips are very short compared to VISEM-Tracking. Our dataset~\cite{vajira_thambawita_2022_7293726} contains $7\times$ more annotated video frames. In addition, VISEM-Tracking contains $2.3\times$ more annotated objects compared to SVIA.

VISEM-Tracking offers annotated bounding boxes and sperm tracking information, making it more valuable for training supervised ML models than the original VISEM dataset~\cite{haugen2019visemDataset}, which lacks these annotations. This additional data enables a variety of research possibilities in both biology (e.g., comparing with CASA tracking) and computer science (e.g., object tracking, integrating clinical and tracking data). Unlike other datasets, VISEM-Tracking's motility features facilitate sperm identification within video sequences, resulting in a richer and more detailed dataset that supports novel research directions. Potential applications include sperm tracking, classifying spermatozoa based on motility, and analyzing movement patterns. To the best of our knowledge, this is the first open dataset of its kind.

\section*{Methods}

The videos for this dataset were originally obtained to study overweight and obesity in the context of male reproductive function~\cite{andersen2016fatty, boivin2007international}. In the study, male participants aged 18 years or older were recruited between 2008 and 2013 from the normal population. Further details on the recruitment can be found in~\cite{haugen2019visemDataset}. The study was approved by the Regional Committee for Medical and Health Research Ethics, South East, Norway (REK number: 2008/3957). All participants provided written informed consent and agreed to the publication of the data. The original project was finished in December 2017, and all data was fully anonymized. 

The samples to be recorded were placed on a heated microscope stage (37\degree C) and examined under a $400\times$ magnification using an Olympus CX31 microscope. The videos were recorded by a microscope-mounted UEye UI-2210C camera made by IDS Imaging Development Systems in Germany. According to the WHO recommendations \cite{world2010laboratory}  light microscope equipped with phase-contrast optics is necessary for all examinations of unstained preparations of fresh semen. The videos are saved as AVI files. Motility assessment was performed based on the videos following the WHO recommendations~\cite{world2010laboratory}.

\begin{figure}[!ht]
    \centering
    \includegraphics[width=\textwidth]{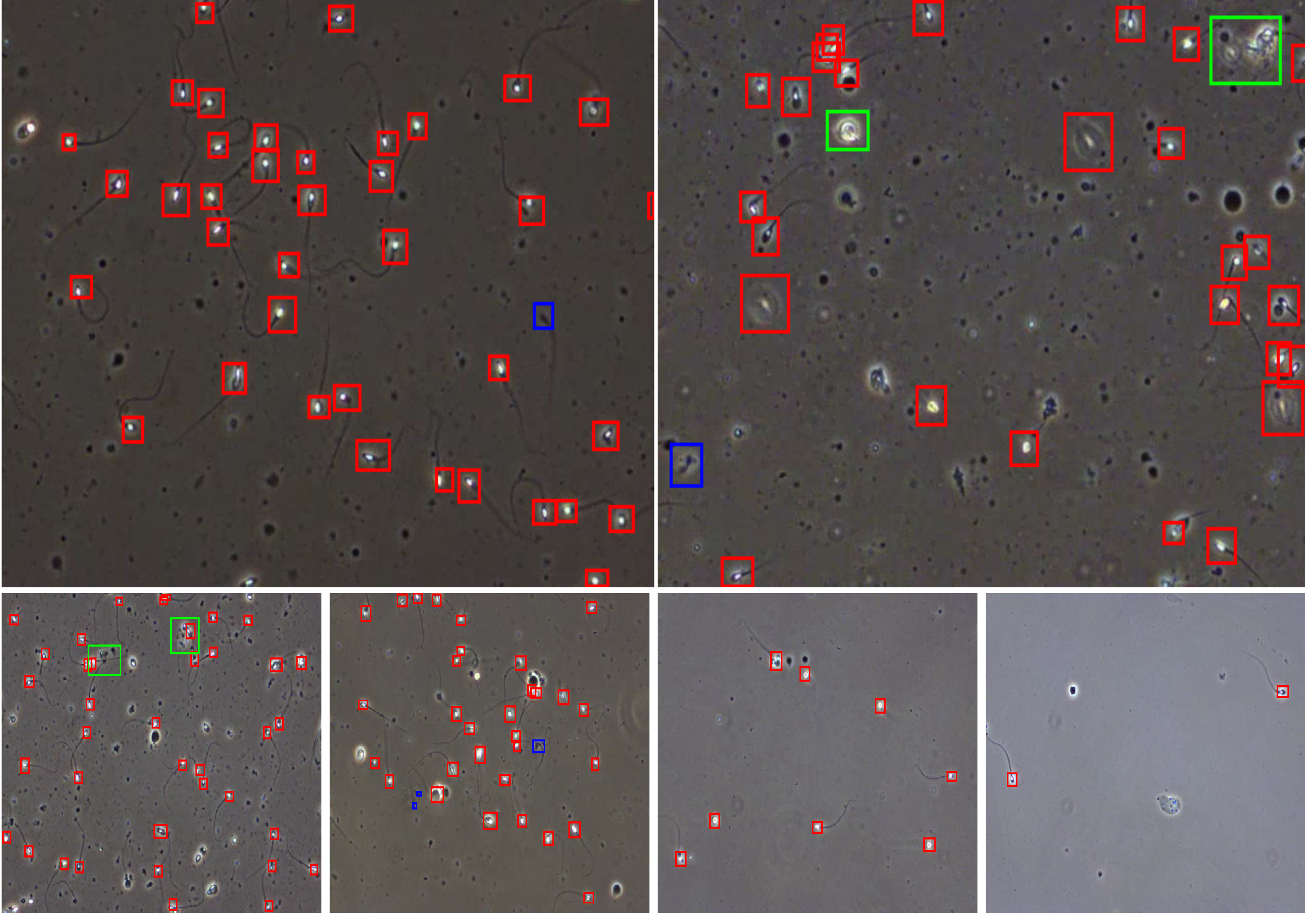}
    \caption{Video frames of wet semen preparations with corresponding bounding boxes. Top: large images showing different classes of bounding boxes, red - sperm, green - sperm cluster, and blue - small or pinhead sperm. Bottom: presenting different sperm concentration levels from high to low (from left to right, respectively).}
    \label{figure:Example_Images_With_BB}
    \vspace{-10pt}
\end{figure}

The bounding box annotation was performed by data scientists in close collaboration with researchers in the field of male reproduction. The data scientists labeled each video using the tool LabelBox (\url{https://labelbox.com}), which was then verified by the three biologists to ensure that the annotations were correct. Moreover, in addition to the per sperm tracking annotation, we also provide additional labels per spermatozoa, which are: `normal sperm', `pinhead', and `cluster'. The pinhead  category consists of spermatozoa with abnormally small black heads within the view of the microscope. The cluster category consists of several spermatozoa grouped together. Sample annotations are presented in Figure~\ref{figure:Example_Images_With_BB}. The red boxes represent normal spermatozoa cells which constitute the majority of this dataset and are also biologically most relevant. The green boxes represent sperm clusters where few spermatozoa cells are clustered together, making it hard to annotate sperm cells separately. The blue color boxes represent small or pinhead spermatozoa which are smaller than normal spermatozoa and have very small heads compared to a normal sperm head. 

\section*{Data Records}
VISEM-Tracking is available at Zenodo (\url{https://zenodo.org/record/7293726})\cite{vajira_thambawita_2022_7293726} and the license for the data is Creative Commons Attribution 4.0 International (CC BY 4.0). This dataset contains mainly \numberofvideos\space  videos (collected from $20$ different patients), each with a fixed duration of $30$ seconds with the corresponding annotated bounding boxes. 
The \numberofvideos\space  were chosen based on how different they are to all the videos in the dataset in order to obtain as many diverse tracking samples as possible.
Since each video from the original dataset lasts for more than $30$ seconds we also provide, in addition to the annotated video clips, the remaining video as \numberofsplitsrest \space ($30$ seconds) video clips for the $20$ annotated videos and \numberofsplitsunlabelled\space ($30$ seconds) video clips for all unlabelled videos of the VISEM dataset~\cite{haugen2019visemDataset} that were not used to provide tracking information. This was done to make it easy to use for future studies that aim to explore more advanced methods such as semi- or self-supervised learning~\cite{van2020survey}.

\begin{figure}[!t]
    \centering
    \includegraphics{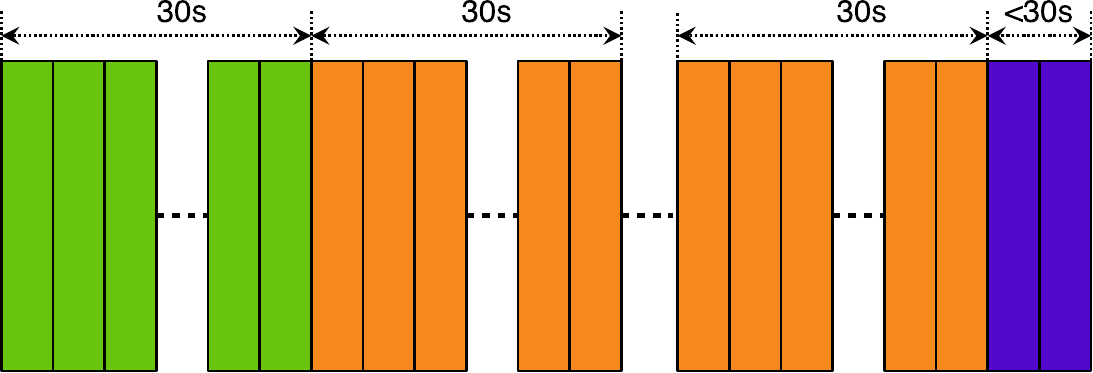}
    \caption{Splitting videos into $30$ seconds clips. \textbf{Green} color represents the split used to manually annotate sperms using bounding boxes. \textbf{Orange} color represents the rest of $30s$ splits included in unlabeled dataset. \textbf{Purple} color section represents the last part of a video which does not have $30s$ long clip. Therefore, we do not include these endings in our dataset to maintain the consistency of $30s$ clips.}
    \label{fig:splitting_video}
\end{figure}

A length of 30 seconds was chosen to make it easier to annotate and process the video files. These videos can also be used for a possible extension of the tracking data in the future. The splitting process of the long videos is presented in Figure~\ref{fig:splitting_video}. More details about the dataset itself are summarized in Table~\ref{tab:data_statistics}.

\begin{table}[!t]
    \centering
    \caption{Summary of quantitative information about the VISEM-Tracking dataset.}
    \begin{tabular}{lc}
        \toprule
        Description & \# Count \\
        \midrule
         \#annotated 30s video clips & $20$ \\
         Frames per second (FPS) per video & $45-50$ \\
         \# of annotated frames & $29,196$ \\
         \# Frames with at least one sperm & $28,974$ \\
         \# Frames with at least one cluster & $10,199$ \\
         \# Frames with at least one small or pinhead sperm & $13,532$ \\
         \# bounding boxes & $656,334$ \\
         \# classes & $3$ (sperm-0, cluster-1, small or pinhead-2) \\
         \# unique sperms (with tracking IDs) & $1,121$ \\
         \# unique clusters (with tracking IDs) & $20$ \\
         \# unique small or pinheads (with tracking IDs) & $35$ \\
         \# unlabeled 30s video clips & $336$ \\
         \# remaining 30s video clips from the $20$ annotated videos & $166$ \\ 
        \bottomrule
    \end{tabular}
    
    \label{tab:data_statistics}
\end{table}

\begin{figure}[!t]
    \centering
    \includegraphics{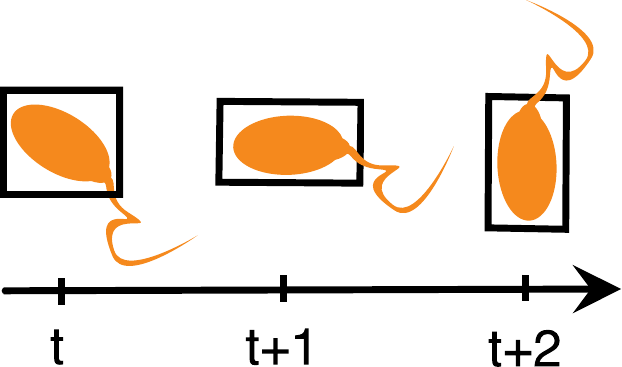}
    \caption{Changing bounding box area over time for the same sperm head.}
    \label{Figure:Sperm_BB_Shapes}
\end{figure}

The folder containing annotated videos has 20 sub-folders with annotations of each video. Each folder of videos has a folder containing extracted frames of the video, a folder containing bounding box labels of each frame, and a folder containing bounding box labels and the corresponding tracking identifiers. In addition to these, a complete video file (.mp4) is provided in the same folder. All bounding box coordinates are given using the YOLO~\cite{glenn_jocher_2022_7002879} format. The folder containing bounding box details with tracking identifiers has `.txt` files with unique tracking ids to identify individual spermatozoa throughout the video. It is worth noting that the area of the bounding boxes of the same sperm changes over time depending on its position and movement in the videos, as depicted in Figure~\ref{Figure:Sperm_BB_Shapes}. Moreover, the text files contain class labels, 0: normal sperm, 1: sperm clusters, and 2: small or pinhead sperm. Additionally, the \emph{sperm\_counts\_per\_frame.csv} file provides per frame sperm count, cluster count, small\_or\_pinhead count. 

\begin{figure}[!t]
    
    \includegraphics[width=\textwidth]{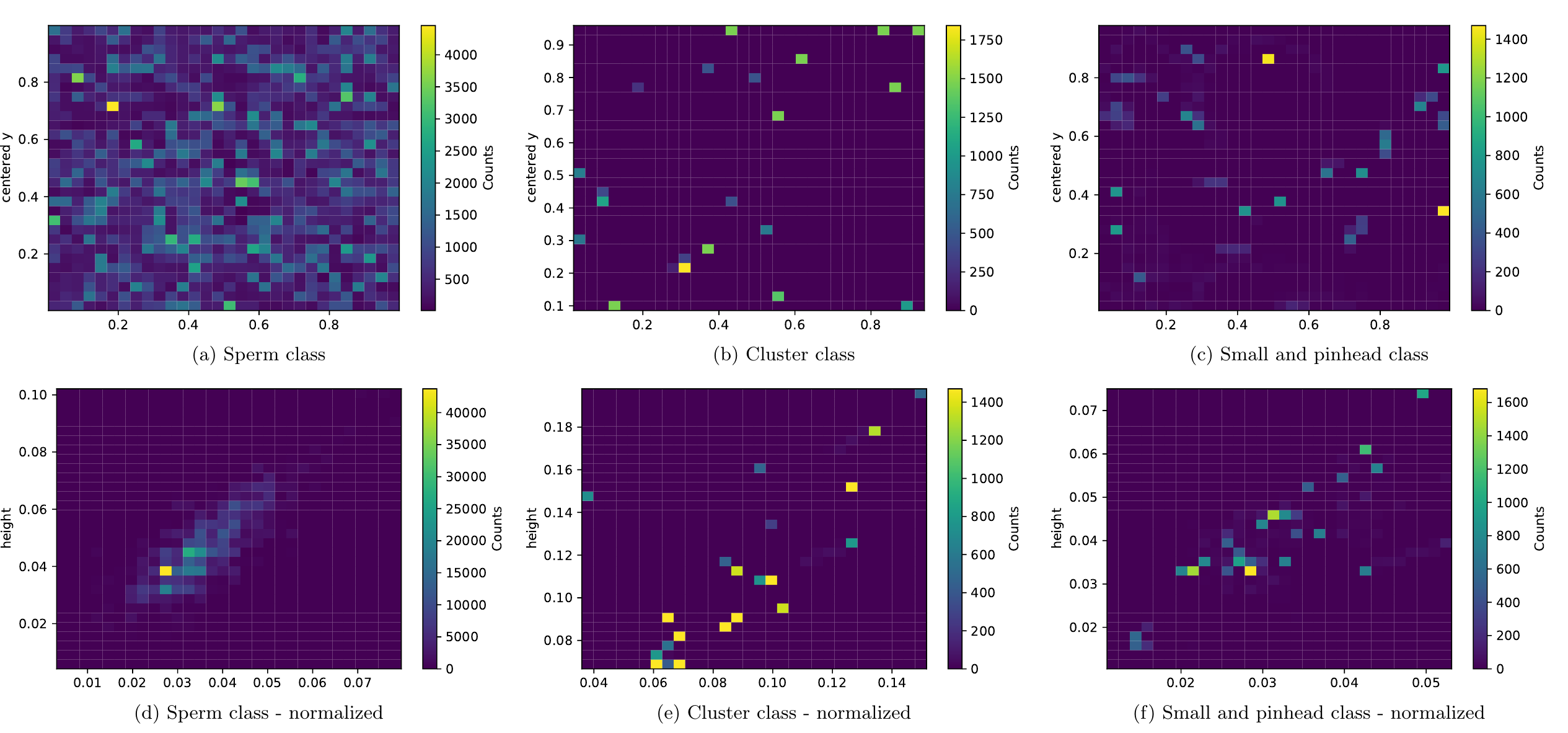}
    \caption{Statistics about bounding box coordinates and area. (a) - 2D histogram on the center coordinates of the bounding boxes of the sperm class. (b) - 2D histogram on the center coordinates of the bounding boxes of the cluster class. (c) - 2D histogram on the center coordinates of the bounding boxes of the small and pinhead class. (d) - 2D histogram on the height and width (normalized values) of the bounding boxes of sperm class. (e) - 2D histogram of the height and width (normalized values) of the bounding boxes of the cluster class. (f) - 2D histogram on the height and width (normalized values) of the bounding boxes of the small or pinhead class.}
    \label{fig:bb_histogram}
\end{figure}

In most of the labeled videos, each frame contains bounding box information ($1,470$ frames on average per video). The video titled \textit{video\_23} has $174$ frames without spermatozoa. Furthermore, some videos are recorded at different frame rates (videos \textit{video\_35} and \textit{video\_52} have $1,440$ total frames, and video \textit{video\_82} has $1,500$ total frames). The distribution of the bounding boxes is reflected in Figure~\ref{fig:bb_histogram}a, and the 2D histogram of the height and width of the bounding boxes is shown in Figure~\ref{fig:bb_histogram}b. Figure~\ref{fig:bb_histogram}a shows that the bounding boxes tend to be evenly distributed across the video frames, with a higher concentration of bounding boxes in the upper left of the video frames. According to Figure~\ref{fig:bb_histogram}b, the variation in bounding box size is quite small.

In addition to the bounding box details, several \emph{.csv} files taken from the VISEM dataset~\cite{haugen2019visemDataset}, are provided with additional information. These files include general information about participants (\emph{participant\_related\_data\_Train.csv}),  the standard semen analysis results (\emph{semen\_analysis\_data\_Train.csv}), serum levels of sex hormones (\emph{sex\_hormones\_Train.csv}: measured from blood samples), serum levels of the fatty acids in the phospholipids (\emph{fatty\_acids\_serum\_Train.csv}: measured from blood samples), fatty acid levels of spermatozoa (\emph{fatty\_acids\_spermatoza\_Train.csv}). The summary of the content of these files is listed in Table~\ref{tbl:csv_summary}.

\begin{table}[!ht]
\renewcommand{\arraystretch}{1.5}
\caption{Summary of content of CSV files included in the VISEM-Tracking dataset.}
\begin{tabular}{lp{11.6cm}}
\toprule
File name &
  File headers \\
\midrule
participant\_related\_data\_Train.csv &
  ID, Abstinence time(days), Body mass index (kg/$m^2$), Age (years) \\
semen\_analysis\_data\_Train.csv &
  ID, Sperm concentration (x$10^6$/mL), Total sperm count (x$10^6$), Ejaculate volume (mL), Sperm vitality (\%), Normal spermatozoa (\%), Head defects (\%), Midpiece and neck defects (\%), Tail defects (\%), Cytoplasmic droplet (\%), Teratozoospermia index, Progressive motility (\%), Non-progressive sperm motility (\%), Immotile sperm (\%), High DNA stainability, HDS (\%), DNA fragmentation index, DFI (\%) \\
sex\_hormones\_Train.csv &
  ID, Seminal plasma anti-Müllerian hormone (AMH) (pmol/L), Serum total testosterone (nmol/L), Serum oestradiol (nmol/L), Serum sex hormone-binding globulin, SHBG (nmol/L), Serum follicle-stimulating hormone, FSH (IU/L), Serum Luteinizing hormone, LH (IU/L), Serum inhibin B (ng/L), Serum anti-Müllerian hormone, AMH (pmol/L) \\
fatty\_acids\_serum\_Train.csv &
  ID, Serum C14:0 (myristic acid), Serum C16:0 (palmitic acid), Serum C16:1 (palmitoleic acid), Serum C18:0 (stearic acid), Serum C18:1 n-9 (oleic acid), Serum total C18:1, Serum C18:2 n-6 (linoleic acid, LA), Serum C18:3 n-6 (gamma-linoleic acid, GLA), Serum C20:1 n-9, Serum C20:2 n-6, Serum C20:3 n-6, Serum C20:4 n-6, Serum C20:5 n-3 (eicosapentaenoic acid, EPA), Serum C22:5 n-3 (docosapentaenoic acid, DPA), Serum C22:6 n-3 (docosahexaenoic acid, DHA) \\
fatty\_acids\_spermatoza\_Train.csv &
  ID, Sperm C14:0 (myristic acid), Sperm C15:0 (pentadecanoic acid), Sperm C16:0 (palmitic acid), Sperm C16:1 n-7 (palmitoleic acid), Sperm C17:0, Sperm C18:0 (stearic acid), Sperm C18:1 trans n-6 to n-11, Sperm C18:1 n-9 (oleic acid), Sperm C18:1 n-7 to n-11, Sperm C18:2 n-6 (Linoleic acid, LA), Sperm C20:0, Sperm C18:3 n-6 (gamma-linoleic acid, GLA), Sperm C18:3 n-3 (a-linoleic acid, ALA), Sperm C20:1 n-9, Sperm C20:2 n-6, Sperm C22:0, Sperm C20:3 n-6, Sperm C20:4 n-6 and C22:1 n-9 combined, Sperm C20:5 n-3 (eicosapentaenoic acid, EPA), Sperm C24:0, Sperm C24:1 n-9, Sperm C22:5 n-3 (docosapentaenoic acid, DPA), Sperm C22:6,n3 (docosahexaenoic acid, DHA) \\
  \bottomrule
\end{tabular}

\label{tbl:csv_summary}
\end{table}

\section*{Technical Validation}

We divided the $20$ videos into a training dataset of 16 videos and a validation dataset of $4$ videos (video IDs of the validation dataset are provided in the GitHub repository). The training set was used to train baseline \gls{dl} models, and the validation dataset was used to evaluate our \gls{dl} models. YOLOv5~\cite{glenn_jocher_2022_7002879} was selected as the baseline sperm detection \gls{dnn} model. This version of YOLO consists of five different models, namely, YOLOv5n (nano), YOLOv5s (small), YOLOv5m (medium), YOLOv5l (large), and YOLOv5x (XLarge). All models were trained using the training dataset with a number of class parameters of $3$, which include \emph{normal sperm, cluster, and small or pinhead} categories. 

In the training process, we provided extracted frames and the corresponding bounding box details to the YOLOv5 models. We set the image size parameter to $640$, batch size to $16$, and the number of epochs to $300$. All other hyperparameters, such as learning rate, batch size, and optimizer, were kept with default values of YOLOv5 (\url{https://github.com/ultralytics/yolov5}). Furthermore, all experiments were performed on two NVIDIA GeForce RTX $3080$ graphic processing units with a total of $20 GB$ memory ($10GB$ per each GPU) with AMD Ryzen 9 $3950X$ 16-Core Processor.  
The best model was found using the performance on the validation dataset. 

Precision, recall, $mAP\_0.5$, $mAP\_0.5:0.95$, and \emph{fitness value}, as calculated by Jocher et al.~\cite{glenn_jocher_2022_7002879}, were used to measure the performance of different YOLOv5 models. The results are listed in Table~\ref{tab:performance_metrics}, showing that YOLOv5l performs best with a fitness value of $0.0920$. The fitness value presented in the table is calculated using the following equation, which is used in the YOLOv5 implementation to compare model performance. 

\begin{align*}
    Fitness\_value = (0.1 \times mAP\_0.5 + 0.9 \times mAP\_0.95)
\end{align*}

\begin{figure}[!t]
   \includegraphics[width=\textwidth]{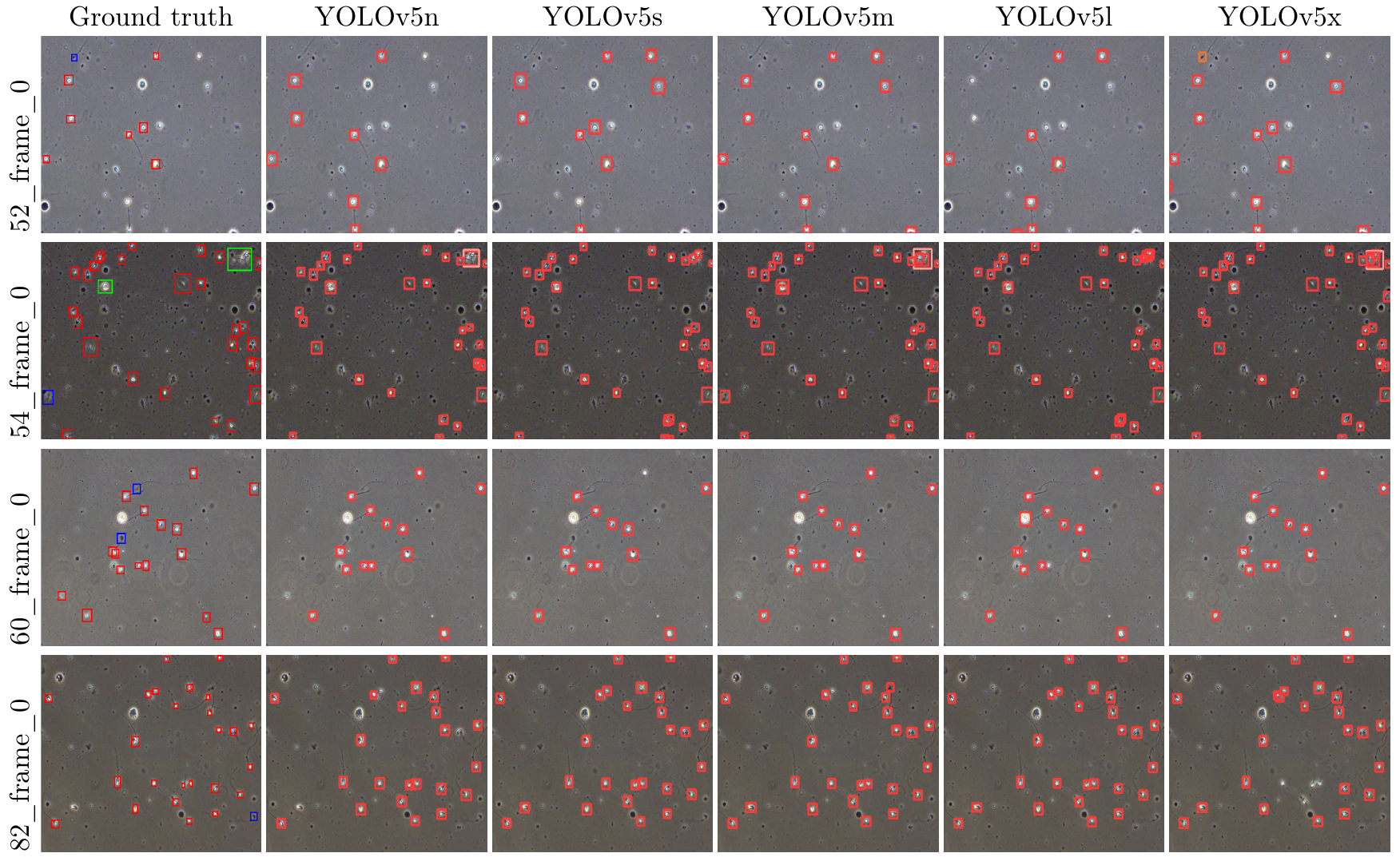} 
    \caption{Predicted bounding boxes from the different models of YOLOv5 for the first frames of the validation data. Video IDs 82, 60, 54 and 52 were used as validation videos.}
    \label{fig:bb_predictions}
\end{figure}

Samples for visual comparisons of predictions from the five models are shown in Figure \ref{fig:bb_predictions}. These predictions are from the first frame of the selected four validation videos.

\begin{table}[!t]
    \centering
    \caption{Different evaluation metrics and corresponding values with the five different YOLOv5 models. }
    \begin{tabular}{l l l l l l}
        \toprule
        \textbf{YOLO model} & \textbf{Precision} & \textbf{Recall} & \textbf{mAP\_0.5} & \textbf{mAP\_0.5:0.95} & \textbf{Fitness value} \\
        \midrule
        YOLOv5n & 0.4120 & 0.2380 & 0.2046 & 0.0567 & 0.0715 \\
        YOLOv5s & 0.4292 & 0.2560 & 0.2102 & 0.0703 & 0.0843 \\
        YOLOv5m & 0.5712 & 0.2279 & 0.2216 & 0.0655 & 0.0811 \\
        YOLOv5l & 0.4323 & 0.2550 & 0.2231 & 0.0775 & 0.0920 \\
        YOLOv5x & 0.3093 & 0.2517 & 0.1995 & 0.0630 & 0.0766 \\
        \bottomrule
    \end{tabular}
    \label{tab:performance_metrics}
\end{table}

\section*{Usage Notes}

To the best of our knowledge, this is the first dataset containing long human spermatozoa video clips (30 seconds with 45-50 FPS) that are manually annotated with bounding boxes for each spermatozoon. The performance of our \gls{dl} experiments for detecting spermatozoa shows that the training data provided in this dataset is diverse and can be used to train advanced \gls{dl} models. 

The data enables different future research directions. For example, it can be used to prepare more labeled data using strategies like semi-supervised learning. Researchers can use the labeled data to train a \gls{dl} model (such as YOLOv5) and predict bounding boxes for the unlabeled data. Then, those pseudo-labeled data can be passed to the experts in the domain to verify them. This method can make the data annotation process easier and produce accurate labeled datasets faster than manual annotations. 

Sperm tracking is necessary to determine sperm dynamics and motility levels. We provide tracking IDs to identify the same spermatozoa throughout the video. Using this data, one can train sperm tracking algorithms, and the results of the tracking algorithms can help to identify different biomedical relevant parameters such as velocity and kinematics. Additionally, it is difficult to determine which spermatozoa in a semen sample have the highest motility, which is of clinical importance. The dataset can be used to train such algorithms for finding spermatozoa with the highest motility. 

In addition to the sperm tracking annotations, we also provide additional metadata for the sperm samples. Using this data, researchers can train models that combine the metadata with the tracking information to obtain more accurate predictions of, for example, motility levels. 

There is also a growing interest in exploring synthetic data to address data deficiencies and timely and costly data annotation problems in the medical domain~\cite{thambawita2021deepsynthbody}. Researchers can use the dataset to train deep generative models~\cite{ho2020denoising, goodfellow2020generative} to generate synthetic data, which then can be used to train other \gls{ml} models and achieve better generalizable performance. Furthermore, one can train conditional deep generative models~\cite{mirza2014conditional, sinha2021d2c} to generate synthetic sperm data with the corresponding ground truth (bounding boxes) using our dataset to overcome the costly problem of getting annotated data.  

Another hot topic in AI and medicine is simulating biological organs or creating digital twins. The dataset can, for example, be used to extract features of sperm motility to simulate spermatozoa and their behaviors. Simulations of spermatozoa can potentially lead to more accurate models than current solutions in the field.

\section*{Code availability}

The code repository with the scripts of data preparations and technical validations (models and pre-trained checkpoints) is available at \url{https://github.com/simulamet-host/visem-tracking}. The original YOLOv5 code is available at \url{https://github.com/ultralytics/yolov5}. 




\section*{Acknowledgements} 

The research presented in this paper has benefited from the Experimental Infrastructure for Exploration of Exascale Computing (eX3), which is financially supported by the Research Council of Norway under contract 270053.

\section*{Author contributions statement}
VT, SAH, and MAR made the conception and design of the work. VT, SAH, AS, HLH, and MAR prepared and annotated data. JA, OW, and TBH reviewed and verified data annotations. VT conceived and conducted the deep learning experiment(s). VT, SAH, AS,TN, and MAR analysed the data and results. VT, AS, SAH, TH, PH, and MAR prepared the draft of the work. All authors reviewed and revised the manuscript.

\section*{Competing interests} 
None of the authors has any competing interests.

\end{document}